\documentclass{article}

    \PassOptionsToPackage{numbers, compress}{natbib}

\usepackage[preprint]{neurips_2023}

\usepackage{listings}
\lstdefinestyle{instructformat}{
  basicstyle=\ttfamily\small,
  frame=single,
  breaklines=true,
  keywordstyle=\color{blue},
  keywords={User, Assistant},
}

\usepackage[utf8]{inputenc} 
\usepackage[T1]{fontenc}    
\usepackage[pagebackref=true,breaklinks=true,colorlinks,bookmarks=false, allcolors=blue]{hyperref}
\usepackage[pdftex]{graphicx}
\usepackage{url}            
\usepackage{booktabs}       
\usepackage{amsfonts}       
\usepackage{nicefrac}       
\usepackage{microtype}      
\usepackage{xcolor}         
\usepackage{epsfig}
\usepackage{caption}
\usepackage{subcaption}
\usepackage{amsmath}
\usepackage{multirow}
\usepackage{array}
\usepackage{pifont}
\usepackage{booktabs}
\usepackage{wrapfig}
\usepackage{hhline}
\usepackage{color, colortbl}
\usepackage{bm}
\usepackage{marvosym}
\newcommand{\tableCellHeight}{1}
\newcommand{\tabstyle}[1]{
  \setlength{\tabcolsep}{#1}
  \renewcommand{\arraystretch}{\tableCellHeight}
  \centering
  \small
}
\definecolor{forestgreen}{rgb}{0.133, 0.545, 0.133}
\definecolor{yellowyellow}{rgb}{0.133, 0.545, 0.133}
\definecolor{COLOR_MEAN}{HTML}{f0f0f0}
\definecolor{ROW_COLOR}{HTML}{C9F7F4}
\definecolor{greendot}{HTML}{06d6a0}
\definecolor{magneta}{HTML}{FE6D73}
\definecolor{dark_green}{HTML}{17C3B2}
\definecolor{wrong_color}{HTML}{FFB29B}

\usepackage[capitalize]{cleveref}
\crefname{section}{Sec.}{Secs.}
\Crefname{section}{Section}{Sections}
\Crefname{table}{Table}{Tables}
\crefname{table}{Tab.}{Tabs.}
\title{OtterHD: A High-Resolution Multi-modality Model}

\author{
  Bo Li\textsuperscript{*}\quad Peiyuan Zhang\textsuperscript{*}  \\ \textbf{Jingkang Yang}\textsuperscript{$\dagger$}\quad\textbf{Yuanhan Zhang}\textsuperscript{$\dagger$}\quad\textbf{Fanyi Pu}\textsuperscript{$\dagger$}\quad\textbf{Ziwei Liu}\textsuperscript{\Letter}\\
  S-Lab, Nanyang Technological University, Singapore \\
{\tt\small\{libo0013, peiyuan.zhang, ziwei.liu\}@ntu.edu.sg}\\
{\tt\small\url{https://github.com/Luodian/Otter}} \\
{\tt\small\url{https://huggingface.co/datasets/Otter-AI/MagnifierBench}}
}

\begin{document}
\maketitle
\renewcommand{\thefootnote}{\fnsymbol{footnote}}
\footnotetext[1]{Equal contribution, \textsuperscript{$\dagger$}Equal appreciation on assistance, \textsuperscript{\Letter}Corresponding author.}
\renewcommand*{\thefootnote}{\arabic{footnote}}
\begin{abstract}

In this paper, we present \textbf{OtterHD-8B}, an innovative multimodal model evolved from Fuyu-8B, specifically engineered to interpret high-resolution visual inputs with granular precision. Unlike conventional models that are constrained by fixed-size vision encoders, OtterHD-8B boasts the ability to handle flexible input dimensions, ensuring its versatility across various inference requirements. 
Alongside this model, we introduce \textbf{MagnifierBench}, an evaluation framework designed to scrutinize models' ability to discern minute details and spatial relationships of small objects. 
Our comparative analysis reveals that while current leading models falter on this benchmark, OtterHD-8B, particularly when directly processing high-resolution inputs, outperforms its counterparts by a substantial margin.
The findings illuminate the structural variances in visual information processing among different models and the influence that the vision encoders' pre-training resolution disparities have on model effectiveness within such benchmarks. 
Our study highlights the critical role of flexibility and high-resolution input capabilities in large multimodal models and also exemplifies the potential inherent in the Fuyu architecture's simplicity for handling complex visual data.
\begin{figure*}[htp]
    \centering
    \includegraphics[width=0.5\textwidth]{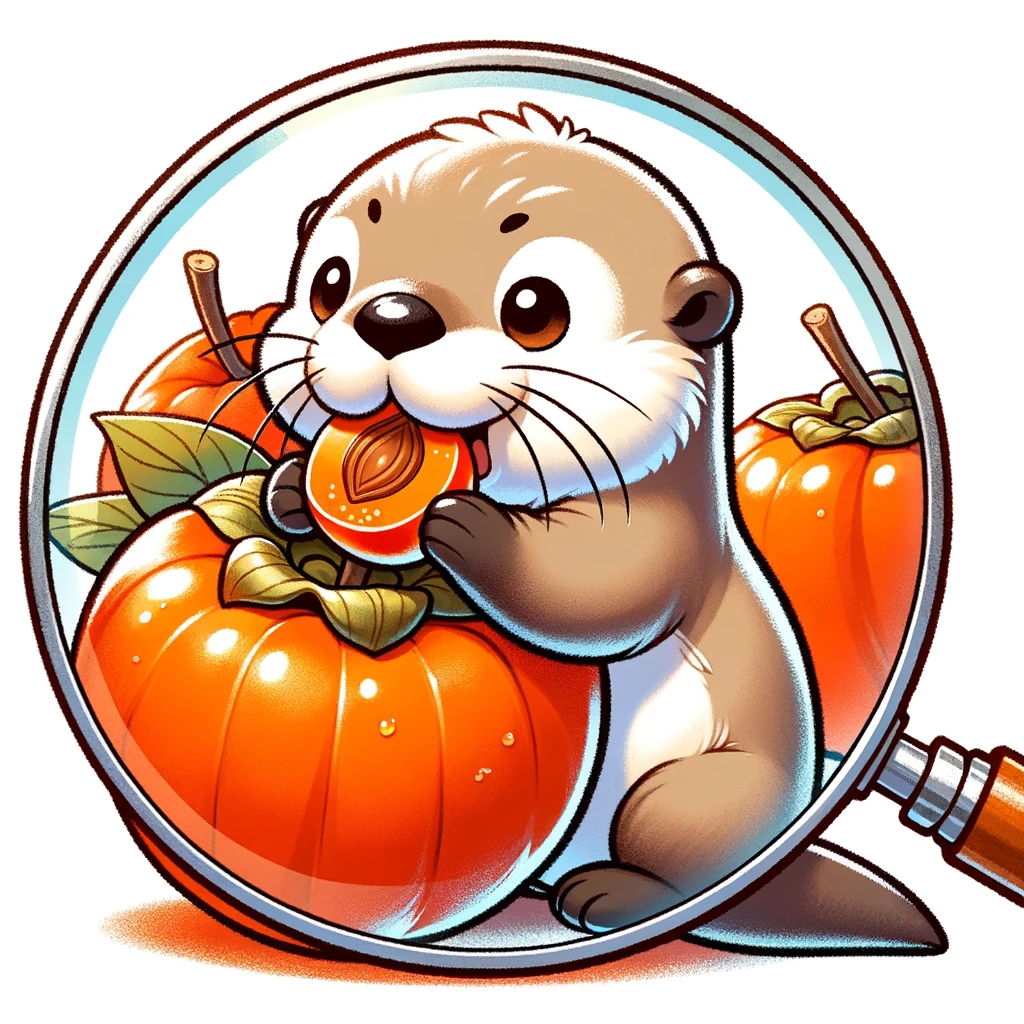}
    \label{fig:teaser}
\end{figure*}
\end{abstract}

\section{Introduction \& Motivation}

The remarkable success achieved by scaling language models~\cite{raffel2023exploring, gpt3, chowdhery2022palm, touvron2023llama} has ignited interest in a similar expansion of Large Multi-modality Models (LMMs)~\cite{llava,open_flamingo,li2023otter,dai2023instructblip,laurenccon2023obelisc}. Interestingly, most recent investigations into LMMs have predominantly centered on enlarging text decoders. For instance, the Llava~\cite{llava} and Qwen-VL~\cite{Qwen-VL} series release models with different sizes of the integrated language models, yet maintaining a consistent vision encoder and input resolution. There has been a relative paucity of efforts directed at amplifying the image component of LMMs. The PaLI series~\cite{chen2022pali, chen2023pali, chen2023pali3} stands out as one of the few research focusing on scaling up the vision encoder.  They also explored increasing input resolution and evaluating the model on fine-grained OCR tasks. Such studies underscore that the concurrent scaling of both vision and language components yield enhanced performance across a broad array of tasks.

The current trend in Large Multimodal Models (LMMs) tends to hinge on the dual-tower architecture, which is composed of a vision encoder, a language decoder, and a bridging mechanism. The vision encoder, exemplified by models such as ViT~\cite{dosovitskiy2021image} and CLIP~\cite{clip}, typically adheres to fixed resolutions like 224$\times$224 or 336$\times$336 during training. While it is possible to introduce higher resolution images during a fine-tuning phase, as demonstrated by models like PaLI, the inference resolution remains unchanged, limiting the model's ability to adapt to varying resolutions and reducing its inference-time flexibility. This rigidity could impede the model's capacity to process and recognize inputs at higher resolutions, despite the vision encoder's prior knowledge about images. Moreover, effectively integrating vision and language models of varying sizes into a cohesive system presents an ongoing and complex challenge for researchers in the field.

Our work is motivated by the Fuyu-8B model~\cite{fuyu-8b}, which elegantly sidesteps these limitations by removing the vision encoder altogether and directly incorporating pixel-level information into the language decoder. The model leverages its native position embeddings to comprehend different image sizes, obviating the need for separate high and low-resolution training stages as seen in the PaLI series.  

Building upon Fuyu, we introduce \textbf{OtterHD-8B}, an advanced instruction-tuned model to handle larger and various image resolutions. OtterHD-8B is open-sourced and the instruction tuning process is specifically designed to accommodate a wide range of image resolutions up to 1024$\times$1024 pixels. Such elasticity allows users to choose the input resolution given their inference budget and task nature.  We evaluate OtterHD on a broad range of benchmarks, including \textbf{MagnifierBench}: a novel benchmark we developed that focuses on evaluating LMMs' capacity to detect minute details in high-resolution images. 
The images in MagnifierBench showcase intricate scenes densely populated with small objects, primarily found in first-person videos of household activities. The dataset compilation process required annotators to meticulously zoom in and concentrate on these diminutive objects, which take up roughly 1\% of the image size.
In our evaluation, we observed that conventional fixed-resolution models demonstrate limited efficacy on this benchmark, yielding accuracy akin to random guessing. In contrast, OtterHD, when provided with high-resolution input, significantly surpasses its counterparts. This study emphasizes the critical importance of adaptable, high-resolution inputs for LMMs and highlights the strengths of Fuyu's simple architectural design.
Our contributions can be summarized as follows:
\begin{itemize}
    \setlength{\itemsep}{0pt}
    \setlength{\parsep}{0pt}
    \setlength{\parskip}{0pt}
    \item We present \textbf{OtterHD-8B}, a novel model based on the Fuyu-8B architecture, optimized for varying input resolutions. Our empirical evaluations suggest that the model exhibits state-of-the-art performance across multiple tasks when instruction-tuned with higher resolutions.
    \item We introduce \textbf{MagnifierBench}, a unique benchmark focused on assessing the capabilities of modern LMMs in recognizing minute attributes and inter-object relationships within large images.
\end{itemize}

\begin{figure*}[htp]
    \centering
    \includegraphics[width=0.99\textwidth]{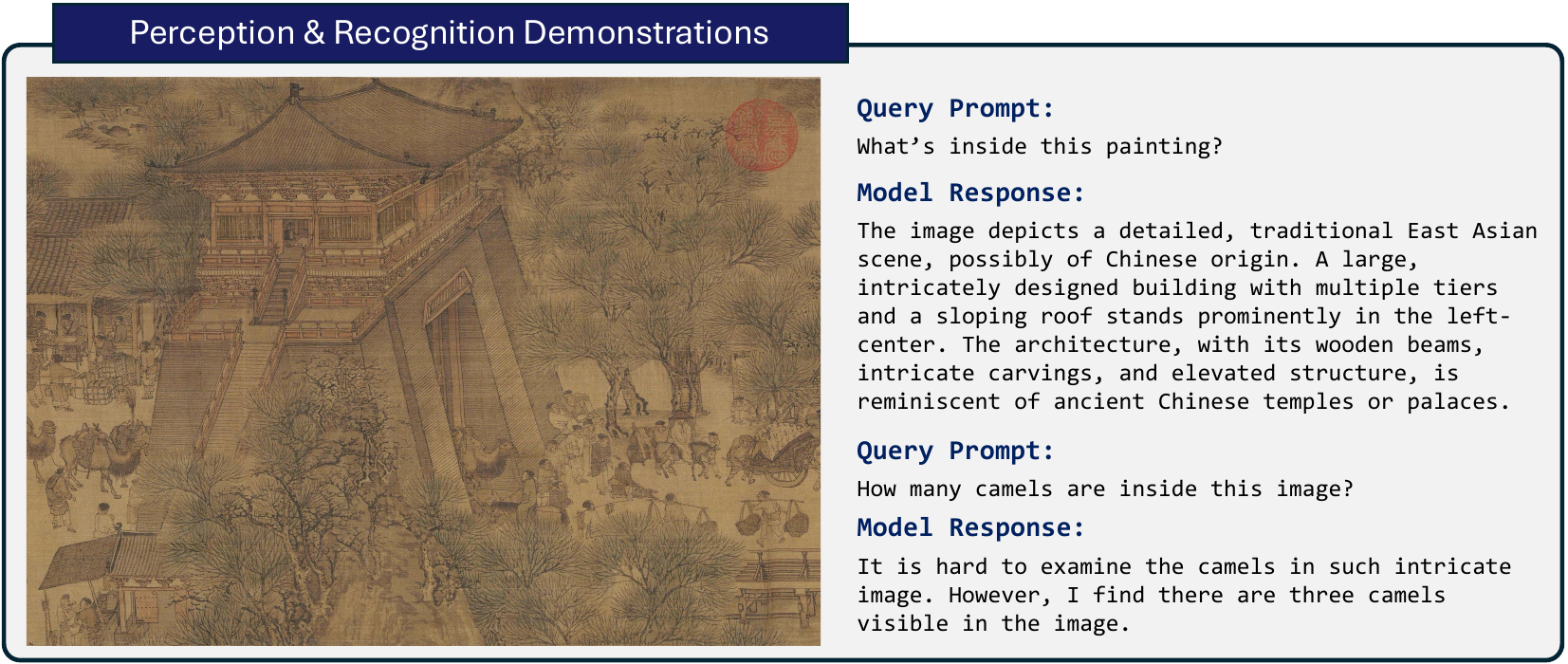}
    \caption{Perception and recognition demonstration of OtterHD-8B. The image is a traditional Chinese painting from the Song Dynasty, \textit{Along the River During the Qingming Festival}. This is a part of the entire artwork, with a resolution of 2466$\times$1766.}
    \label{fig:demo0}
    \vspace{-2mm}
\end{figure*}


\begin{table}[htp]
\tabstyle{3pt}
\renewcommand{\arraystretch}{1.3}
\caption[Caption for LOF]{Performance comparison of OtterHD-8B with prominent open-sourced LMMs, detailing instruction/response data pairs, training, and evaluation resolutions. The term \textit{Dynamic} refers to training with varied resolutions as elaborated in~\cref{subsec:insight}. The term \textit{Original} indicates evaluations using each image's resolution without any resizing operation, whereas other models undergo evaluations with images resized to a consistent square resolution at denoted in \texttt{Eval Res}. Details on metrics are provided in~\cref{subsec:benchmark}.}
\label{tab:main_results}
\resizebox{0.99\textwidth}{!}{%
\begin{tabular}{@{}c|c|c|c|cc|cc|c|c|c|c@{}}
\toprule
\multirow{2}{*}{\textbf{Models}} & \multirow{2}{*}{\textbf{I/R Pairs}} & \multirow{2}{*}{\textbf{Train Res.}} & \multirow{2}{*}{\textbf{Eval Res.}} & \multicolumn{2}{c|}{\textbf{MagBench}} & \multicolumn{2}{c|}{\textbf{MME}$^{1}$} & \multirow{2}{*}{\textbf{POPE}} & \multirow{2}{*}{\textbf{MM-V}} & \multirow{2}{*}{\textbf{MMB}} & \multirow{2}{*}{\textbf{M-Vista}} \\
 &  &  &  & \textbf{Multi.} & \textbf{FF.} & \textbf{Cog.} & \textbf{Percep.} &  &  &  &  \\ \midrule
Idefics-9B$_{\text{instruct}}$~\cite{laurenccon2023obelisc} & 1M & 224 & 224 & 20.8 & 13.4 & 187.9 & 1165.0 & 74.6 & 23.7 & 45.5 & 19.8 \\
Otter-9B~\cite{li2023otter} & 150K & 224 & 224 & 25.7 & 15.8 & 306.4 & 1292.3 & 72.5 & 24.7 & 48.3 & 19.7 \\
InstructBLIP-7B~\cite{dai2023instructblip} & 1.2M & 224 & 224 & 5.6 & 15.2 & - & - & - & 26.2 & 36.0 & - \\
InstructBLIP-13B~\cite{dai2023instructblip} & 1.2M & 224 & 224 & 3.8 & 16.3 & 291.8 & 1212.8 & 78.9 & 25.6 & 33.9 & \textbf{25.3} \\
LLaVA-7B$_{1.5}$~\cite{liu2023improved} & 3.6M$^{2}$ & 336 & 336 & 26.8 & 24.7 & - & \textbf{1510.7} & 85.9 & \textbf{30.5} & \underline{59.5} & - \\
Qwen-VL-7B$_{\text{chat}}$~\cite{Qwen-VL} & 1.4B & 448 & 448 & 14.5 & 15.9 & \textbf{360.7} & 1487.5 & - & - & \textbf{61.8} & - \\ \midrule
Fuyu-8B~\cite{fuyu-8b} & - & - & \textit{Original} & 29.3 & 15.2 & 237.5 & 728.6 & 74.1 & 21.4 & 10.7 & 20.6 \\ \midrule
\multirow{3}{*}{\textbf{OtterHD-8B}} & \multirow{3}{*}{370K} & 512 & 512 & 33.5 & 31.4 & 289.8 & \underline{1359.3} & \textbf{86.1} & 25.1 & 58.5 & 22.3 \\
&  & 1024 & 1024 & 37.8 & 37.2 & 288.5 & 1313.7 & 81.5 & 19.8 & 53.6 & 17.3 \\ 
&  & \textit{Dynamic} & \textit{Original} & \textbf{42.7} & \textbf{39.9} & \underline{331.4} & 1223.4 & \underline{86.0} & \underline{26.3} & 58.3 & \underline{23.5} \\ \bottomrule
\end{tabular}
}
\end{table}
\section{Unified Architecture for Vision and Language}

In this section, we first introduce the background of Fuyu-8B~\cite{fuyu-8b} and Fuyu's backbone language model, Persimmon-8B~\cite{persimmon-8b}. We then detail our proposed multi-modal instruction-following model, OtterHD.

\subsection{Model Design}
\renewcommand{\thefootnote}{}
\footnotetext[1]{$^{1}$ The metric for MME is reported by scores, while for other benchmarks, by accuracies. $^{2}$ The converted instruction/response pairs in the LLaVA-1.5's 665K data, where they put multiple instruction/response pairs towards one image into one sentence. The conversion is made to align with those used in other models in measuring how many instructions are tuned.}
\renewcommand*{\thefootnote}{}

\paragraph{Perssimon-8B~\cite{persimmon-8b}} Persimmon-8B is a decoder-only transformer with modifications like squared ReLU activation~\cite{so2022primer}, rotary positional encodings~\cite{su2022roformer}, and decoupled input$\backslash$output embeddings. It also includes a layernorm for the Q and K embeddings before attention calculation~\cite{dehghani2023scaling}. The model has a hidden size of 4096, 64 heads, and 36 layers, and has seen 737 billion tokens during training. The released checkpoint has approximately 9.3B parameters, making it slightly larger than Llama-7B~\cite{llama}, and its inference cost is comparable to an 8B parameter model with combined embeddings.
\paragraph{Fuyu-8B~\cite{fuyu-8b}}
Fuyu-8B mirrors Persimmon-8B in its design as a decoder-only transformer tailored to both image and text input without an image encoder. Images are divided into 30 by 30 patches and processed similarly to text using causal attention. These patches are tokenized in a raster-scan order, with a unique "image-newline" character indicating line breaks of each row. The model uses its inherent position embeddings to understand varying image sizes, eliminating the necessity for distinct high and low-resolution training phases like the PaLI series. 

\paragraph{OtterHD-8B}
Our OtterHD-8B is a model instruction-tuned from Fuyu-8B, aiming at examining the impact of increasing resolutions on the performance of downstream tasks. We used the following instruction format and used Fuyu's natively defined \texttt{$\backslash$x04} as the beginning of the answer token. 

\begin{lstlisting}[style=instructformat]
     {image tokens} User:{instruction} Assistant:\x04 {answer} \eos
\end{lstlisting}
Similar to Fuyu-8B, images are first resized to a specified target size and then segmented into patches of size 30x30, with padding applied to the bottom and right edges. For ablation studies and comparative analysis, the target size can be set to a fixed or randomly sampled resolution ranging from 448$\times$448 to 1024$\times$1024, as elaborated in~\cref{sec:exp}. We did not explore image augmentation methods such as random cropping. By scaling up the original image to a larger resolution while maintaining a fixed patch size, the patches effectively capture finer details with a smaller receptive field. Notably, OtterHD represents the first open-source instruction-tuned LMM trained on inputs up to 1024$\times$1024. As demonstrated in~\cref{sec:exp}, it further generalizes to even larger resolutions (\textit{e.g.} 1440$\times$1440) during inference. 

\subsection{Training Details}
\label{subsec:training_details}

In preliminary experiments, we found that the Fuyu model exhibited limitations in responding to specific instructions within certain benchmarks, such as not being able to respond well to option letters and yes or no. This results in the very weak performance on MME~\cite{fu2023mme} and MMBench~\cite{liu2023mmbench}.

To address these shortcomings, we embarked on instruction tuning Fuyu model on our data mixture and used a new instruction template. However, the amount of our instruction tuning training is relatively small compared to state-of-the-art LMMs~\cite{llava,Qwen-VL}, there's a possibility that Fuyu's original capabilities might be compromised to some extent.

\textbf{Data Mixture}
We compiled a total of 370K instruction/response pairs sourced from the following public datasets: LLaVA-Instruct~\cite{liu2023improved}, VQAv2~\cite{antol2015vqa}, GQA~\cite{hudson2019gqa}, OKVQA~\cite{marino2019ok}, OCRVQA~\cite{mishraICDAR19}, A-OKVQA~\cite{schwenk2022okvqa}, COCO-GOI~\cite{liu2022open}, COCO-Caption~\cite{chen2015microsoft}, TextQA~\cite{singh2019towards}, RefCOCO~\cite{yu2016modeling}, COCO-ITM~\cite{li2022blip}, ImageNet~\cite{deng2009imagenet}, and LLaVA-RLHF~\cite{sun2023aligning}. The data mixture and specific prompt strategies are motivated by \texttt{LLaVA-1.5}~\cite{liu2023improved} and \texttt{Idefics-Instruct}~\cite{laurenccon2023obelisc} to achieve better text formatting control. All the datasets were organized into instruction/response pairs, aggregated into a single dataloader and uniformly sampled during the training phase to ensure representational integrity.

On average, each instruction/response pair produces approximately $200$ text tokens and $342$ image tokens including \texttt{|NEWLINE|} tokens, when the input resolution is set to $512\times512$. Further details, including the average dimensions of images in each dataset, can be found in~\cref{app:data_mixture}.

\textbf{Implementation \& Optimization}

\begin{figure*}[htp]
    \centering
    \includegraphics[width=0.90\textwidth]{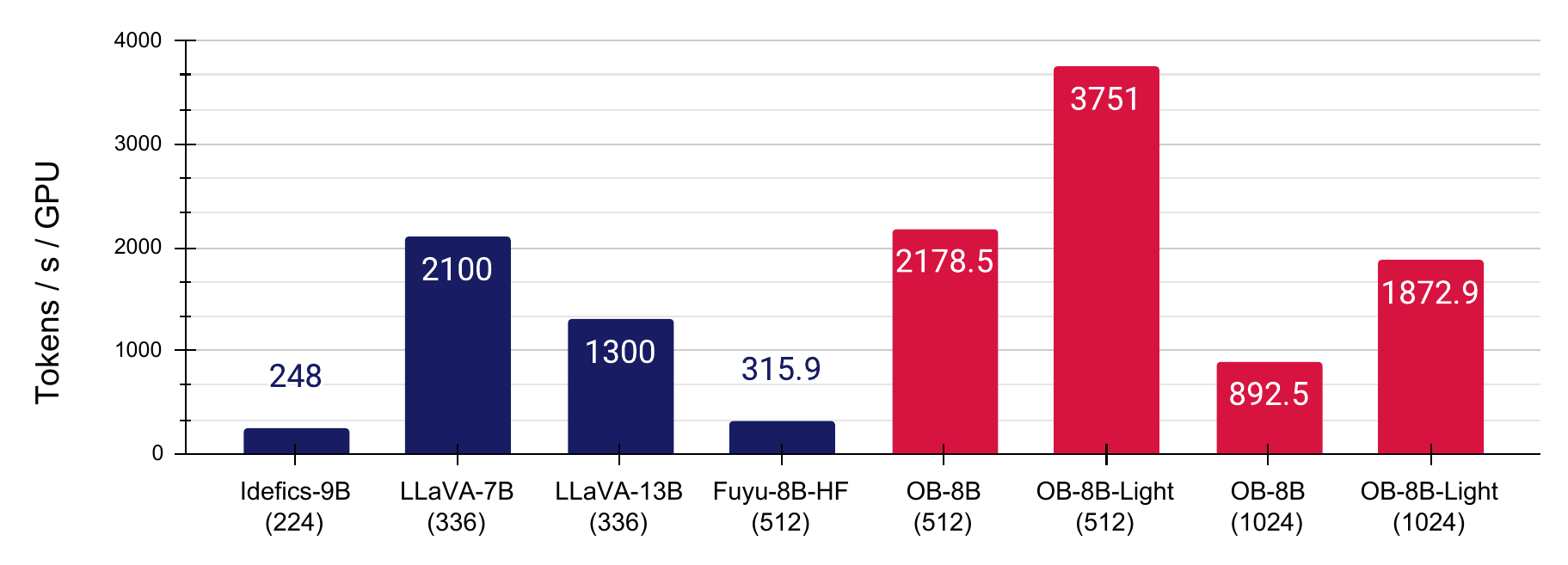}
    \caption{Comparative assessment of throughput across different models. The training throughput metric, denoted as \textit{tokens per second per GPU}, is determined by recording the values for each batch and subsequently computing the average over a 30-minute duration. The tokens encompasses both image and text tokens.}
    \label{fig:throughput}
\end{figure*}

Our experiments utilize the PyTorch library in conjunction with the HuggingFace transformers~\cite{wolf-etal-2020-transformers} framework. We find that the native HuggingFace implementation of Fuyu-8B is highly unoptimized. We thus augment the modeling code with FlashAttention-2~\cite{dao2023flashattention2} and other fused operators including fused layernorm, fused square ReLU, and fused rotary positional embedding from the FlashAttention repository~\cite{dao2023flashattention2}. Fuyu's simplified architecture facilitates us to do this in a fairly convenient way. As illustrated in~\cref{fig:throughput}, the modifications substantially enhance GPU utilization and throughput. 

In the configurations, OB refers to finetuning with full parameters, whereas OB-Light indicates LoRA finetuning with $r=32$ and $\alpha=32$. The targeted modules for modification encompass all attention and linear layers, including the \texttt{head} layer. 

Our implementation permits the completion of full-parameter training within $3$ hours per epoch on $8\times$A100 GPUs. Additionally, LoRA finetuning requires just $1$ hour per epoch. The model is trained with a batch size of 64 using the AdamW optimizer, set with a learning rate of $1\times10^{-5}$ and a weight decay of $0.1$. A discussion on full-parameters and LoRA tuning are provided in~\cref{app:lora_finetune} and more details are provided in the~\cref{app:hparam}.

\section{MagnifierBench}

\begin{figure*}[tp]
    \centering
    \includegraphics[width=0.98\textwidth]{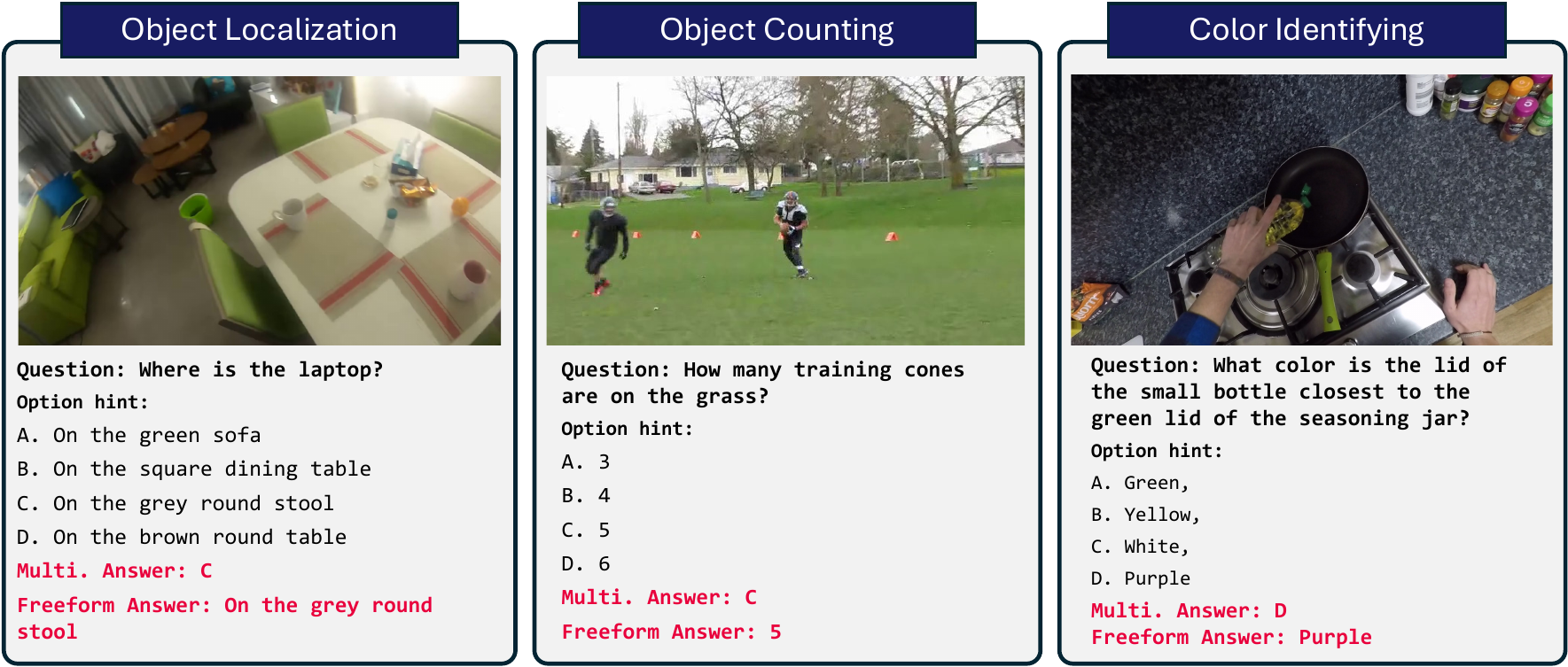}
    \caption{Sample demonstrations of the three types of questions in the \textbf{MagnifierBench}. Each question is associated with two types of the questions and answers. The resolutions are 1080$\times$1920 pixels for both left and right images, while the central image has 640$\times$480 pixels.}
    \label{fig:zibench_examples}
\end{figure*}

The human visual system can naturally perceive the details of small objects within a wide field of view, but current benchmarks for testing LMMs have not specifically focused on assessing this ability. This may be because the input sizes of mainstream Vision-Language models are constrained to relatively small resolutions. With the advent of the Fuyu and OtterHD models, we can, for the first time, extend the input resolution to a much larger range. Therefore, there is an urgent need for a benchmark that can test the ability to discern the details of small objects in high-resolution input images. In this paper, we introduce the MagnifierBench to fill this gap.

\subsection{Construction Details}
The images of MagnifierBench are sourced from the Panoptic Scene Graph Generation (PVSG) dataset~\cite{yang2023panoptic}, which consists of video data featuring a plethora of complex scenes cluttered with miscellaneous objects, especially in first-person videos of household chores. To utilize the PVSG dataset,  our annotation team was instructed to first scrutinize the videos to identify distinctive complex frames, characterized by the presence of numerous small objects. A small square, equivalent to 1\% of the image size, was placed beside each video to assist annotators in gauging the scale of the small items. Once suitable frames were identified and noted, the annotators' next task was to develop question-answer pairs of those minute objects.  As depicted in Figure \ref{fig:zibench_examples}, each question comes with the query itself and four potential answers. Our dataset offers two answer formats: multiple-choice options and freeform responses. In the subsequent post-annotation phase, our authorial team meticulously reviewed each question-answer entry in the dataset. We eliminated any questions that referred to excessively large objects or those that could be easily answered with common sense knowledge. For instance, questions about the color of a remote control were removed, as most remotes are black, making it an easy guess and excluding colors like red or yellow.

The resulting MagnifierBench dataset aggregates 283 question-answer (QA) pairs derived from 166 images sourced from the PVSG dataset~\cite{yang2023panoptic}. Specifically, the collection comprises 172 QA pairs from 108 images in EpicKitchen~\cite{epic}, 80 QAs from 38 images in Ego4D~\cite{ego4d}, and 31 QA pairs from 20 images in VidOR~\cite{vidor}. The typical resolution of images from EpicKitchen and Ego4D is 1920 $\times$ 1080 pixels, while VidOR is usually 640 $\times$ 480 pixels. 

Figure~\ref{fig:zibench_examples} shows the examples from the MagnifierBench. The types of questions crafted include identification, numerical, color-related questions, and beyond. We emphasized the importance of creating distractor answers that are plausibly confusing, yet ensuring that the correct answer remains unambiguous and singular, as illustrated in the accompanying figure. A crucial criterion for this dataset is that the questions are intricate enough to necessitate the annotator to be in close proximity to the screen, zoom in, and be in full-screen mode on a computer in order to accurately respond. The dataset is readily accessible and can be downloaded from \href{https://huggingface.co/datasets/Otter-AI/MagnifierBench}{\texttt{Otter-AI/MagnifierBench}}.

\subsection{Evaluation Methods}
\label{eval-method}
Recent LMMs are increasingly tuned for generating extended responses in conversational settings as opposed to short answers. Building on previous evaluation techniques~\cite{liu2023mmbench}, we split our assessment into two separate protocols, each designed to quantify the model's performance differently.

\textbf{Multiple Choice:} In this protocol, the model faces a question accompanied by several answer options. To steer the model towards responding with a single letter (\textit{e.g.} A, B, C), we prepend the instruction \textit{Answer with the option letter from the given choices directly} as hint before question to prompt models respond in desired format. In this scenario, only answers that exactly match the correct choice are deemed accurate.

\textbf{Free-Form Answering:} Providing multiple-choice options can simplify the task, as a random guess has a 25\% chance of being correct. Furthermore, it does not reflect the real scenarios faced by chat assistants, where users typically do not present the model with predefined options. To eliminate this potential bias, we also present questions to the model in a straightforward, open-ended manner without any hinting options. We utilize GPT-4 to evaluate the model's response against the benchmark answer, yielding a \texttt{yes} or \texttt{no} verdict for accuracy calculation. The prompt templates for GPT-4, along with sample responses from both assessment types, can be found in~\cref{app:zib_eval}.



\section{Experiments \& Analysis}
\label{sec:exp}
In this section, we analyze the performance of OtterHD evaluated on both our proposed MagnifierBench and several established LMM benchmarks, as outlined in~\cref{subsec:benchmark}. Next, in Section~\cref{subsec:insight}, we share insights garnered during the experimental process. Finally, we demonstrate how OtterHD's performance compares with state-of-the-art models in various real-world scenarios in~\cref{subsec:demos}.

\subsection{Benchmark Evaluation Results}
\label{subsec:benchmark}
In Table~\ref{tab:main_results}, we present a comprehensive comparison between OtterHD-8B and other state-of-the-art LMMs across a variety of benchmarks. We present performance in accuracy on benchmarks including POPE~\cite{li2023evaluating}, MM-Vet~\cite{yu2023mm}, MMBench~\cite{liu2023mmbench}, MathVista~\cite{lu2023mathvista}, and our newly developed MagnifierBench under both the multi-choice protocol and the free-form answering protocol. On MMBench, we report results on \texttt{test set}. For MME~\cite{fu2023mme}, we report the aggregated scores in cognitive and perception to follow its evaluation convention.  We include three different setups for OtterHD: (1) train and test with a fixed resolution at either $512^2$ or $1024^2$. (2) employ a dynamic training approach where images are randomly resized to resolutions from the set [$418^2$, $512^2$, $768^2$, $1024^2$] while testing is conducted at the images' native resolution in the test set. Our findings reveal that while many models achieve high scores on established benchmarks such as MME and POPE, their performance often falls short on our MagnifierBench, demonstrating the necessity of such benchmarks for a more holistic evaluation of LMMs' perceptual ability on fine-grained details. On the other hand, OtterHD-8B showcases outstanding performance on MagnifierBench. Notably, its accuracy improves with higher resolutions. OtterHD-8B also is capable of adjusting to varied image resolutions and aspect ratios in the test set when the training process involves dynamic resizing of images.
Our overall results highlight OtterHD-8B's versatility and superior capability in handling a broad spectrum of tasks and resolutions, making it an exemplary choice for a wide range of multi-modal applications.

\subsection{Empirical Insights}
\label{subsec:insight}

\paragraph{Increasing Resolution and Image-Text Ratios}
\label{para:token_ratio}

\begin{figure*}[tp]
    \centering
    \includegraphics[width=0.90\textwidth]{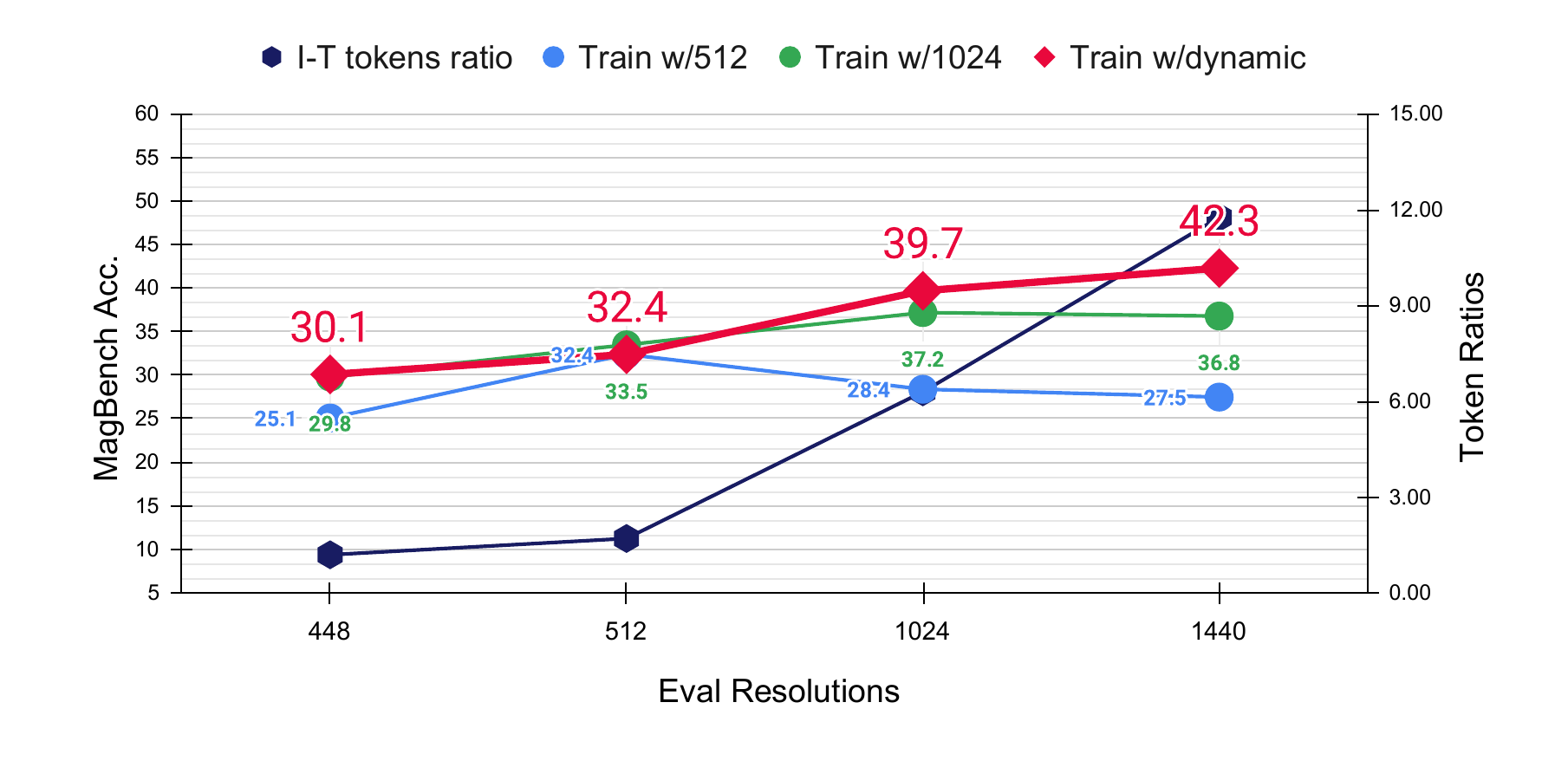}
    \caption{Comparison of OtterHD's performance at different evaluation resolutions. The meanings of \textit{fixed} and \textit{dynamic} are explained in~\cref{para:token_ratio}.}
    \label{fig:token_ratio}
    \vspace{-5mm}
\end{figure*}

To further explore the effect of increasing resolution and OtterHD's ability to generalize to different, potentially larger resolutions, we train Otter8B with fixed or dynamic resolution and present results in \cref{fig:token_ratio}. The $x$-axis suggests that, as the resolution grows during evaluation, more image tokens are sent to the language decoder, offering more details of the image. We compare the performance on MagnifieBench when evaluating across different resolutions under two training strategies. \textit{Fixed} represents using the same resolution to square-resize images during training. \textit{Dynamic} means that images are resized to different dimensions sampled uniformly from $[448, 512, 768, 1024]$ during training. We evaluate the two strategies on various resolutions, including $1440$ to further test if the model can generalize to even larger resolutions. \cref{tab:data_mixture} further shows the image tokens, the image newline tokens, and the average text tokens of MagnificerBench's question-answer pair of each setup. 

\begin{wraptable}{r}{0.3\textwidth}
\tabstyle{5pt}
\renewcommand{\arraystretch}{1.2}\vspace{-5mm}
\caption{Image and text token counts at varying resolutions.}
\label{tab:image_text_ratios}
\resizebox{0.3\textwidth}{!}{%
\begin{tabular}{@{}r|cccc@{}}
\toprule
\textbf{Resolution} & 448 & 512 & 768 & 1024 \\ \midrule
\textbf{Image T.} & 225 & 324 & 676 & 1225 \\ \hline
\textbf{Newline T.} & 15 & 18 & 26 & 35 \\ \hline
\textbf{Text T. (Avg.)} & 200 & 200 & 200 & 200 \\ \bottomrule
\end{tabular}
}
\end{wraptable}

The results reveal that increasing resolution results in better performance on MagnifierBench.  Since the average text tokens remain unchanged, the image-to-text token ratio gradually increases, and it suggests a detailed focus on the image, emphasizing the potential importance of tasks that require detailed visual recognition. This progression accentuates the significance of resolution in LMMs, particularly for tasks necessitating intricate visual entailment. Additionally, the performance variance between the \textit{fixed} and \textit{dynamic} training approaches highlights the benefits of dynamic resizing, especially in preventing overfitting to specific resolutions. The \textit{dynamic} strategy further allows the model to generalize to a larger resolution (1440) not seen during training.

\subsection{Qualitative Demonstrations}
\label{subsec:demos}
\begin{figure*}[htp]
    \centering
    \includegraphics[width=0.95\textwidth]{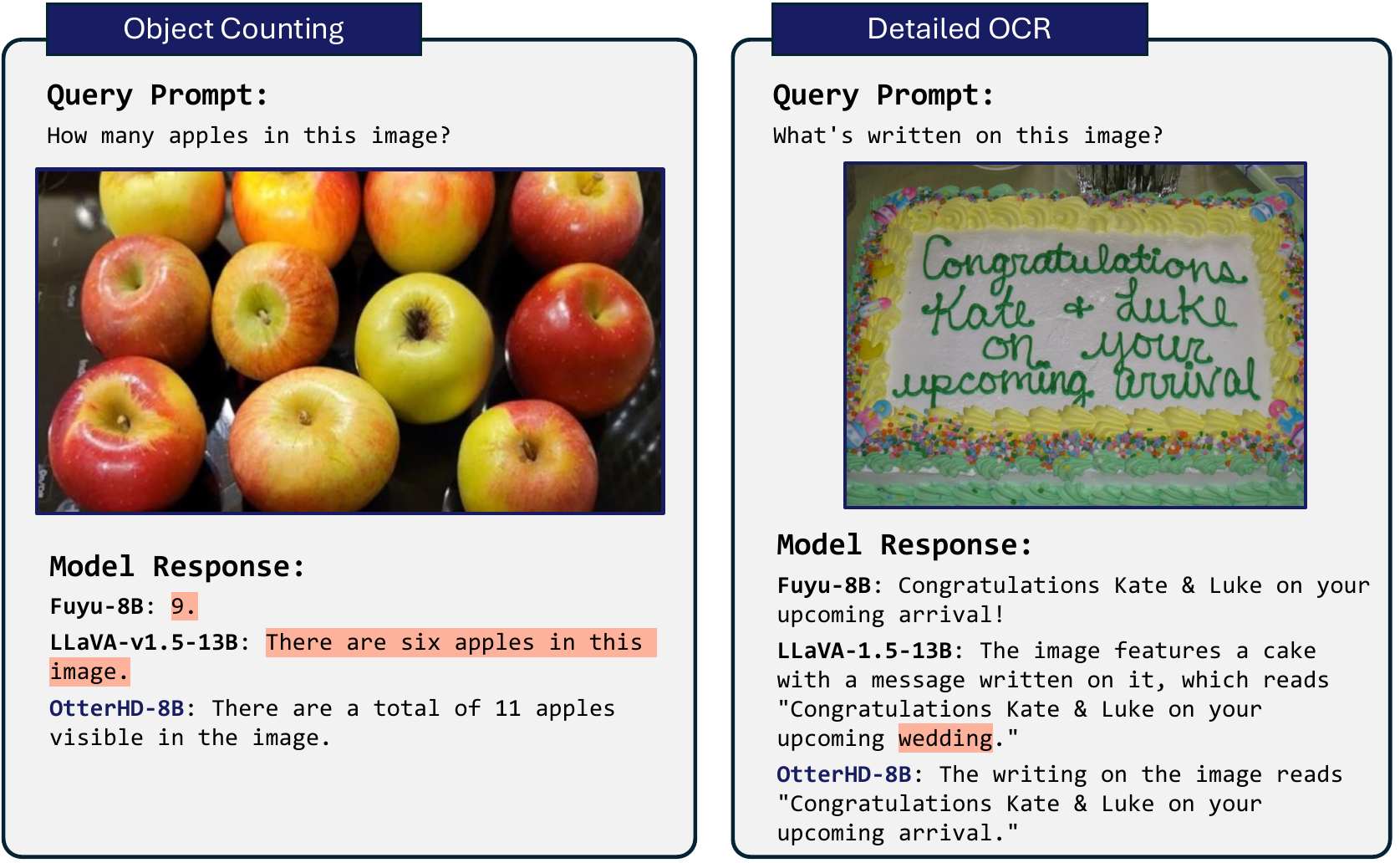}
    \caption{Object counting and detailed scene text understanding abilities comparison between LMMs. \colorbox{wrong_color}{Incorrect parts are noted.}}
    \label{fig:demo1}
\end{figure*}

\begin{figure*}[htp]
    \centering
    \includegraphics[width=0.95\textwidth]{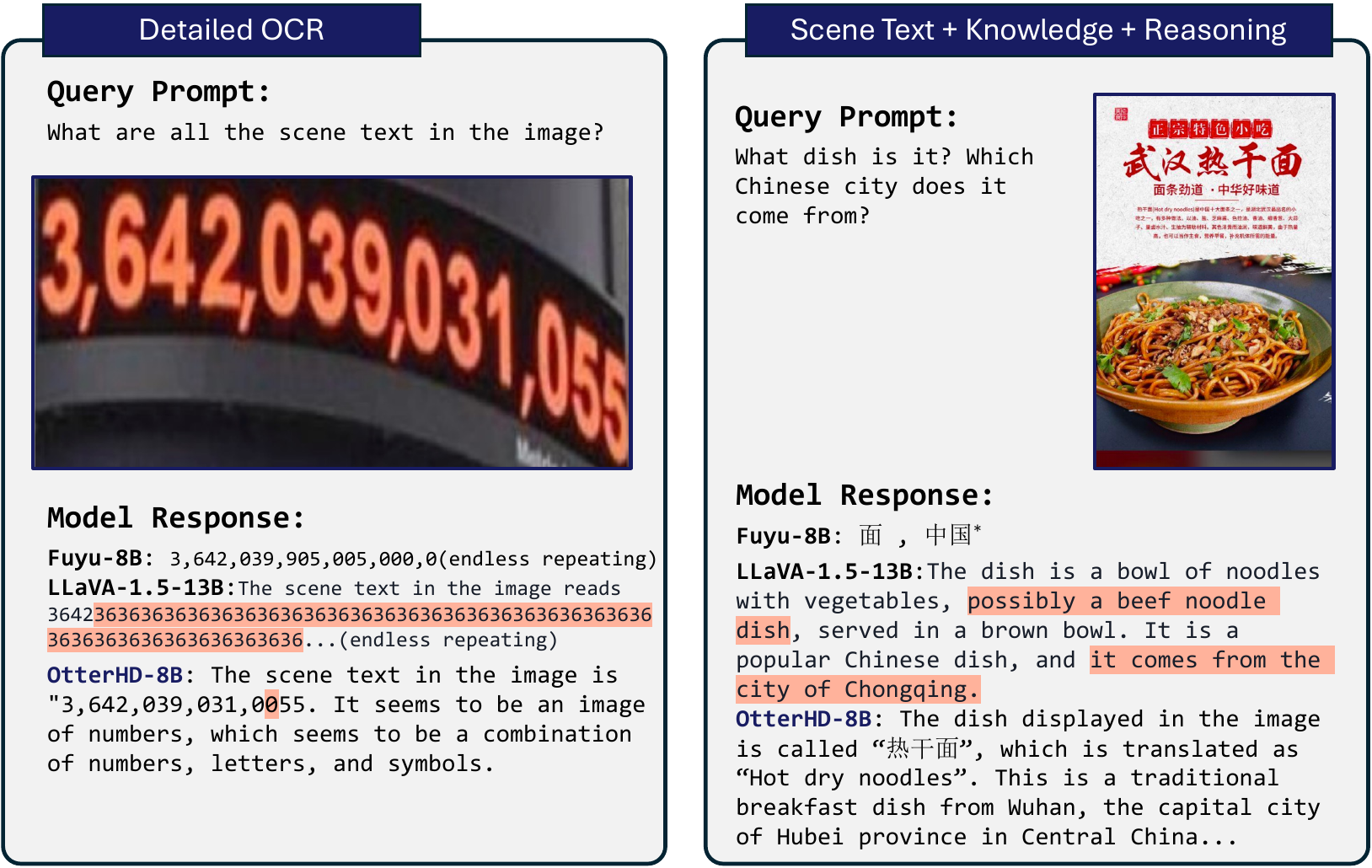}
    \caption{Desktop understanding abilities comparison between LMMs. \colorbox{wrong_color}{Incorrect parts are noted.}}
    \label{fig:demo4}
\end{figure*}

\begin{figure*}[htp]
    \centering
    \includegraphics[width=0.95\textwidth]{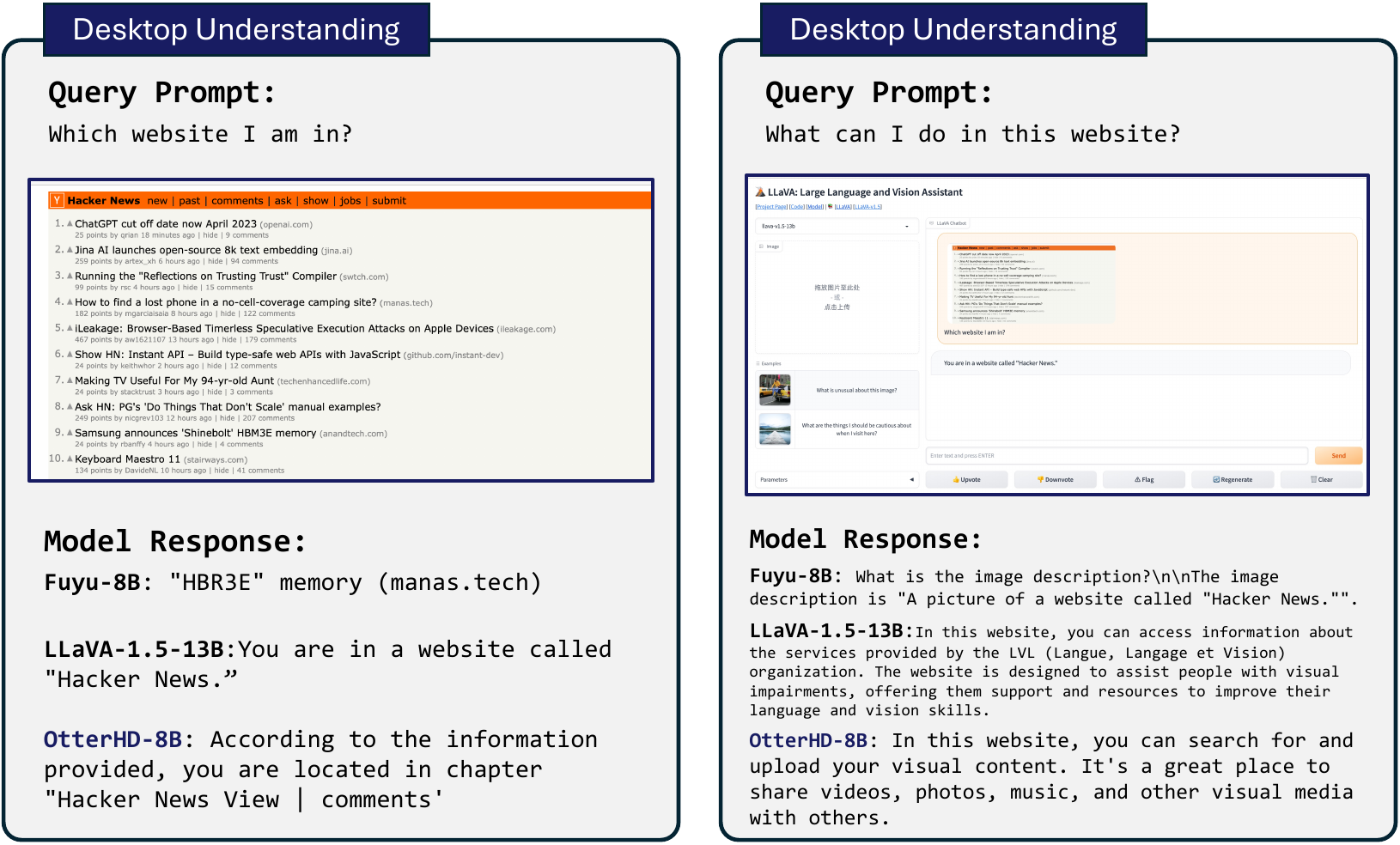}
    \caption{Detailed scene text (desktop oriented) understanding and reasoning abilities comparison between LMMs. \colorbox{wrong_color}{Incorrect parts are noted.}}
    \label{fig:demo2}
\end{figure*}

\begin{figure*}[htp]
    \centering
    \includegraphics[width=0.95\textwidth]{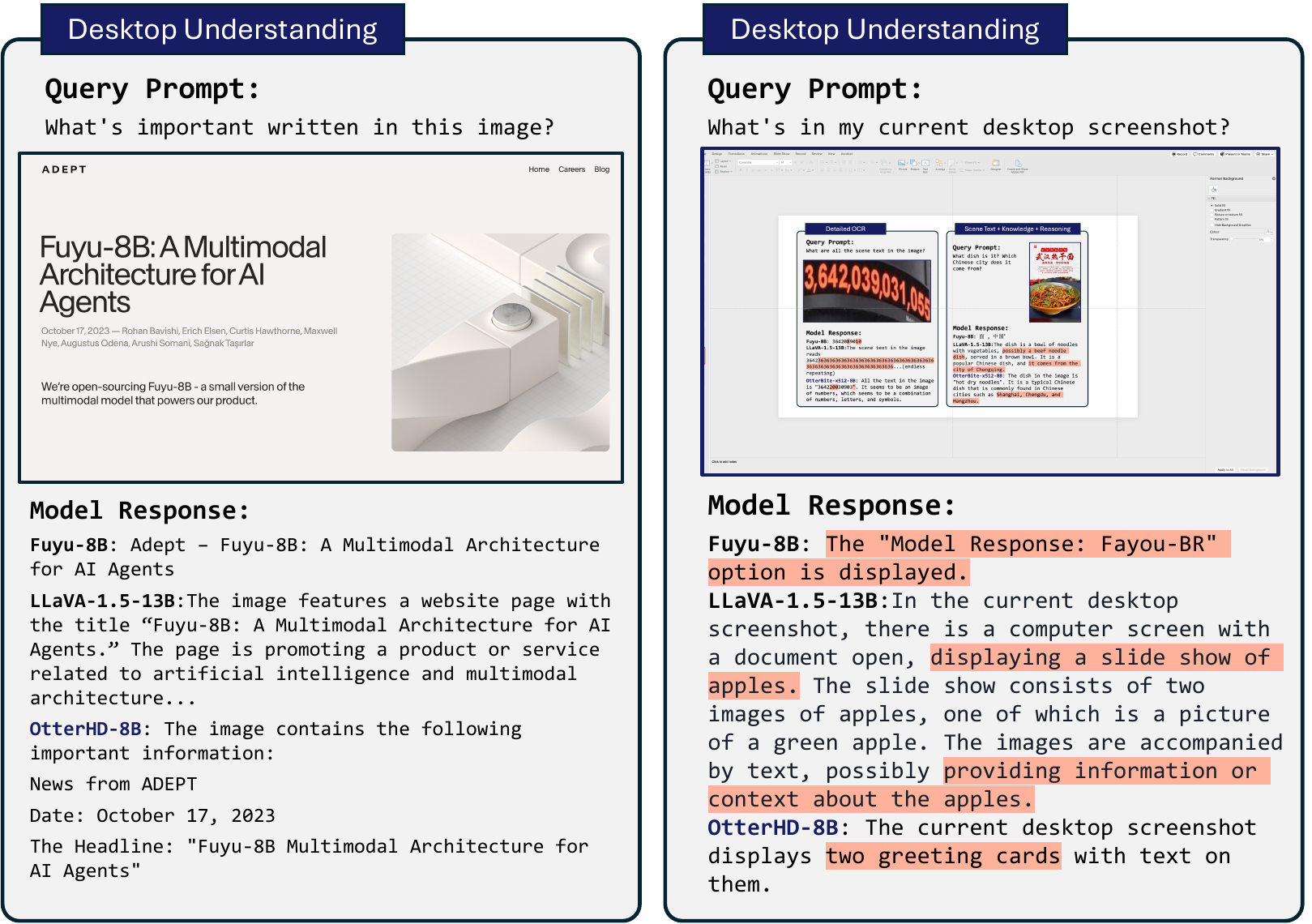}
    \caption{Detailed scene text (desktop oriented) understanding and reasoning abilities comparison between LMMs. \colorbox{wrong_color}{Incorrect parts are noted.}}
    \label{fig:demo3}
\end{figure*}

We bolster our findings with qualitative demonstrations presented in~\cref{fig:demo1,fig:demo2,fig:demo3,fig:demo4}. These illustrations shed light on the performance of OtterHD-8B relative to other LMMs in real-world scenarios, encompassing object counting, detailed scene text comprehension, and screenshot understanding.

\section{Related Work}
\subsection{Large Mulit-modality Models}
The increasing success of large models that focus on a single modality, such as language models~\cite{chatgpt,gpt4,llama,alpaca,vicuna2023} and vision models~\cite{radford2021learning,fang2023eva}, has sparked a recent surge in research exploring combinations of these models. The objective is to integrate models from various modalities into cohesive, end-to-end trainable structures, which are termed Large Multi-modal Models (LMM). As delineated by Zhang et al.~\cite{zhang2023transfer}, the architectures of current LMMs can be segmented into three components: a vision encoder, a projector, and a large language model (LLM). Depending on variations in the VPG and projector setups, the designs of contemporary LMMs can be classified into four types:
\textbf{(1) vision encoder + resampler + cross-gated attention layer}: This category comprises models like Flamingo~\cite{flamingo,open_flamingo} and Otter~\cite{li2023otter}. Significantly, Otter is an enhanced version of OpenFlamingo~\cite{open_flamingo} with optimized instructions. Here, the resampler processes a varying number of image or video features from the vision encoder, producing a fixed number of visual tokens, thus reducing the computational intensity of the vision-text cross-attention. The cross-gated attention layer block is constructed by inserting a freshly initialized cross-attention layer before the frozen self-attention block in the original LLM's cross-attention layer.
\textbf{(2) vision encoder + Q-former + linear layer}: Models like BLIP-2~\cite{li2023blip} are representatives of this configuration, with instructBLIP~\cite{dai2023instructblip} as its instruction-optimized variant. This design omits the intricate \textit{cross-gated attention layer block} found in Flamingo and adopts a streamlined linear layer as the cross-modality projector. Q-former is a small transformer that utilizes a collection of learnable query vectors to glean visual features from the stationary image encoder.
\textbf{(3) vision encoder + linear layer}: LLaVA~\cite{llava} epitomizes this setup. In this configuration, LLaVA retains all vision tokens to prompt the LLM, preserving the entirety of the visual information.
\textbf{(4) linear layer only}: Models in this category, such as Fuyu, operate as basic decoder-only transformers without a specialized vision encoder. In this category, image patches are directly transformeed by a linear layer and projected into the language decoder layers. The advantage of this design lies in its independence from pre-trained vision encoders for information processing. Consequently, the model is not constrained by the fixed resolutions adapted by pre-trained vision encoders, allowing for a more natural adaptation to higher-resolution image inputs. Our OtterHD model also employs this design approach.

\subsection{Benchmarking Detailed Perception}

Grasping intricate visual details, particularly those of smaller objects, is crucial for computer vision models to be effectively applied in real-world scenarios such as autonomous driving and robotics~\cite{liu2021survey, tong2020recent}. However, within the Large Multimodal Models (LMMs) domain, current models and benchmarks have not sufficiently addressed this requirement.
Benchmarks like MME~\cite{fu2023mme}, MMBench~\cite{liu2023mmbench}, and SEED-Bench~\cite{li2023seedbench} do evaluate the perceptual abilities of LMMs, yet they do not adequately focus on the nuanced perception of smaller objects. While tasks related to Optical Character Recognition (OCR)~\cite{mori1999optical, mishraICDAR19,mathew2021docvqa,sidorov2020textcaps} may appear to be suited for evaluating fine-grained details, they are predominantly concerned with text recognition. 
In this work, we underscore the critical need to enhance LMMs’ performance in detailed perception, especially in relation to smaller objects. We emphasize the importance of specialized benchmarks such as MagnifierBench, aiming to close the existing gaps and expand the capabilities of LMMs in terms of perception and understanding.


\section{Conclusion}

In this study, we present the OtterHD-8B model, which builds on the innovative architecture of Fuyu-8B. This model effectively processes images of various resolutions, moving away from the traditional limitation of fixed-resolution inputs seen in most LMMs. Specifically designed for following instructions, OtterHD-8B excels in dealing with high-resolution images. This becomes especially evident when tested against the new MagnifierBench benchmark that is designed to evaluate the capability of LMMs to discern fine details in complex scenes, highlighting the crucial role of resolution flexibility in contemporary LMMs. Our results not only spotlight the promise of Fuyu-like architectures for future studies but also underscore the need for benchmarks like MagnifierBench to rigorously test LLMs' fine-grained perception.
\appendix

\section{Extended Details}



\subsection{Data Mixture \& Average Resolutions}
\label{app:data_mixture}

Table \ref{tab:data_mixture} offers a detailed comparison of the average image resolutions (width and height in pixels) and the number of instruction/response pairs in multiple datasets. This table provides essential insights into the data heterogeneity and scale, serving as a crucial reference for understanding the computational and statistical characteristics of the datasets involved in our model training.

\begin{table}[htp]
\centering
\caption{Summary of average width, height and number of instruction/response pairs across various datasets in our data mixture. The width and height are measured in pixels.}
\label{tab:data_mixture}
\renewcommand{\arraystretch}{1.4}
\resizebox{0.90\textwidth}{!}{%
\begin{tabular}{@{}r|ccccccc@{}}
\toprule
\textbf{Dataset} & \textbf{LLaVA-DD/CR}  & \textbf{VQAv2}  & \textbf{GQA}     & \textbf{OKVQA}    & \textbf{OCRVQA}   & \textbf{A-OKVQA}    & \textbf{COCO-GOI} \\
\midrule
\textbf{Avg. W}  & 577   & 581   & 495   & 617   & 352   & 587   & 586    \\
\textbf{Avg. H} & 481   & 482   & 409   & 448   & 490   & 482   & 476    \\
\textbf{Pairs}       & 53240 & 20000 & 30000 & 18018 & 16354 & 34112 & 20000  \\
\midrule
\textbf{Dataset} & \textbf{COCO-Caption} & \textbf{TextQA} & \textbf{RefCOCO} & \textbf{COCO-ITM} & \textbf{ImageNet} & \textbf{LLaVA-RLHF} & \textbf{Combined} \\
\midrule
\textbf{Avg. W}  & 578   & 950   & 591   & 577   & 469   & 340   & 542    \\
\textbf{Avg. H} & 484   & 811   & 486   & 484   & 387   & 572   & 467    \\
\textbf{Pairs}       & 20000 & 19293 & 20000 & 20000 & 50000 & 50000 & 371017 \\
\bottomrule
\end{tabular}
}
\end{table}

\subsection{Hyperparameters}
\label{app:hparam}

\cref{tab:hparam} provides a comparative overview of the hyperparameters used in two different instruction tuning approaches: LoRA and Full-finetune. This comparison serves to elucidate the computational requirements and settings that yield optimal performance for each approach. However, as the optimal settings may vary based on the computational resources available and the complexity of the problem being addressed.

\begin{table}[htp]
\setlength{\tabcolsep}{3pt}
\renewcommand{\arraystretch}{1.6}
\caption{Comparison of hyperparameter settings between the LoRA and Full-finetune approaches.}
\label{tab:hparam}
\resizebox{0.99\textwidth}{!}{%
\begin{tabular}{@{}r|c|c|c|c|c|c|c|c@{}}
\toprule
\textbf{H-Params} &
  \textbf{Batch Size} &
  \textbf{LR} &
  \textbf{LR Schedule} &
  \textbf{LR Warmup Ratio} &
  \textbf{Epoch} &
  \textbf{Optimizer} &
  \textbf{DeepSpeed} &
  \textbf{Peak Mem. / GPU} \\ \midrule
\textbf{LoRA} &
  128 &
  \multirow{2}{*}{1e-5} &
  \multirow{2}{*}{cosine} &
  \multirow{2}{*}{0.03} &
  6 &
  \multirow{2}{*}{AdamW} &
  \multirow{2}{*}{Zero2} &
  $\sim$70G \\
\textbf{Full-finetune} &
  64 &
   &
   &
   &
  3 &
   &
   &
  $\sim$72G \\ \bottomrule
\end{tabular}
}
\end{table}


\subsection{Full-parameters \textit{vs.} LoRA finetuning}
\label{app:lora_finetune}

\begin{figure*}[htp]
    \centering
    \includegraphics[width=0.98\textwidth]{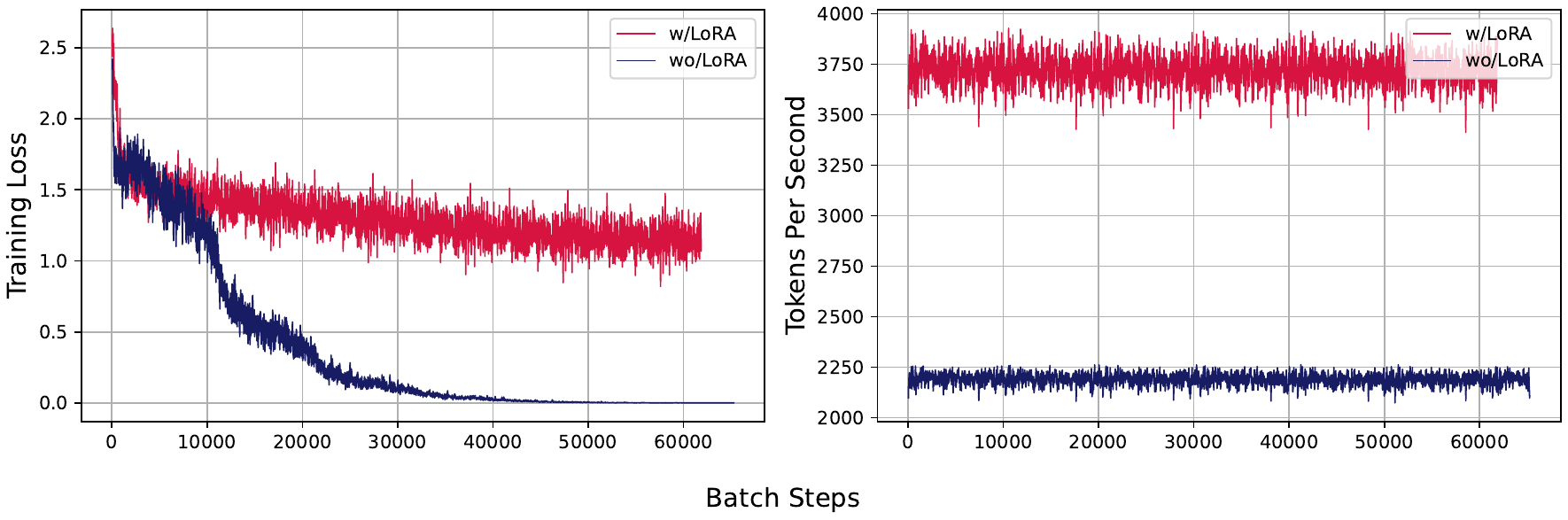}
    \caption{Training loss and token processing efficiency comparison. \textbf{Left:} Training loss trajectories for models with and without Low-Rank Adaptation (LoRA) over batch steps during finetuning. \textbf{Right:} Token processing rates per GPU for models finetuned with full parameters and LoRA.}
    \label{fig:lora}
\end{figure*}

In assessing the efficacy of Low-Rank Adaptation (LoRA) on model performance during finetuning, we observed distinct training behaviors as delineated in Figure \ref{fig:lora}. The left plot of the figure elucidates that integrating LoRA results in a more stable and consistent reduction in training loss over batch steps, indicative of an enhanced learning efficiency as opposed to the conventional full parameter finetuning approach. Furthermore, the right plot in Figure \ref{fig:lora} showcases a significantly higher token processing rate per GPU when utilizing LoRA, highlighting its contribution to improved computational efficiency.


For quantitative performance assessment, Table \ref{tab:performance_comparison} contrasts the outcomes of full-finetuning and LoRA-finetuning techniques. Employing the same training regimen on the LLaVA-Instruct-150K dataset \cite{llava} for a single epoch, the LoRA-SFT approach achieved a notable decrease in the estimated training duration, plummeting from three hours to merely one hour. This remarkable reduction in training time came with only a marginal dip in performance metrics on MagBench and MM-Vet benchmarks. These observations accentuate the practical benefits of LoRA, which offers a compelling tradeoff between efficiency and model performance, making it an attractive alternative for resource-constrained environments or scenarios requiring rapid model iteration.

\begin{table}[htp]
\centering
\setlength{\tabcolsep}{6pt}
\renewcommand{\arraystretch}{1.4}
\caption{Comparison of model performance between the Full-finetune and LoRA-finetune methods. The table displays the number of instruction/response pairs (I/R Pairs), the number of epochs used during training, the estimated time to completion, and performance on the MagBench and MM-Vet benchmarks.}
\label{tab:performance_comparison}
\resizebox{0.60\textwidth}{!}{%
\begin{tabular}{r|c|c|cc}
\toprule
\textbf{Models} & \textbf{I-T Pairs} & \textbf{Epoch} & \textbf{MagBench} & \textbf{MM-Vet} \\ \midrule
\textbf{Fuyu-8B} & - & - & 29.3 & 21.4 \\ \hline
\textbf{Full-params.} & 150K & 1 & \textbf{32.6} & \textbf{24.0} \\
\textbf{LoRA} & 150K & 1 & \underline{29.9} & \underline{22.1} \\ \bottomrule
\end{tabular}
}
\end{table}

The insights gleaned from our experiments suggest that the utilization of LoRA enables researchers and practitioners to significantly cut down on computational resources and time, which are often considerable bottlenecks in the finetuning of large language models. While there is an observable tradeoff in terms of a slight reduction in benchmark performance, the decrease is relatively small when weighed against the benefits of reduced training time and computational load. This balance positions LoRA-finetuning as a strategically viable approach, particularly when quick model deployment is prioritized, and computational efficiency is of paramount importance. 

\subsection{MagnifierBench Evaluation Details}
\label{app:zib_eval}

\begin{figure*}[htp]
    \centering
    \includegraphics[width=0.85\textwidth]{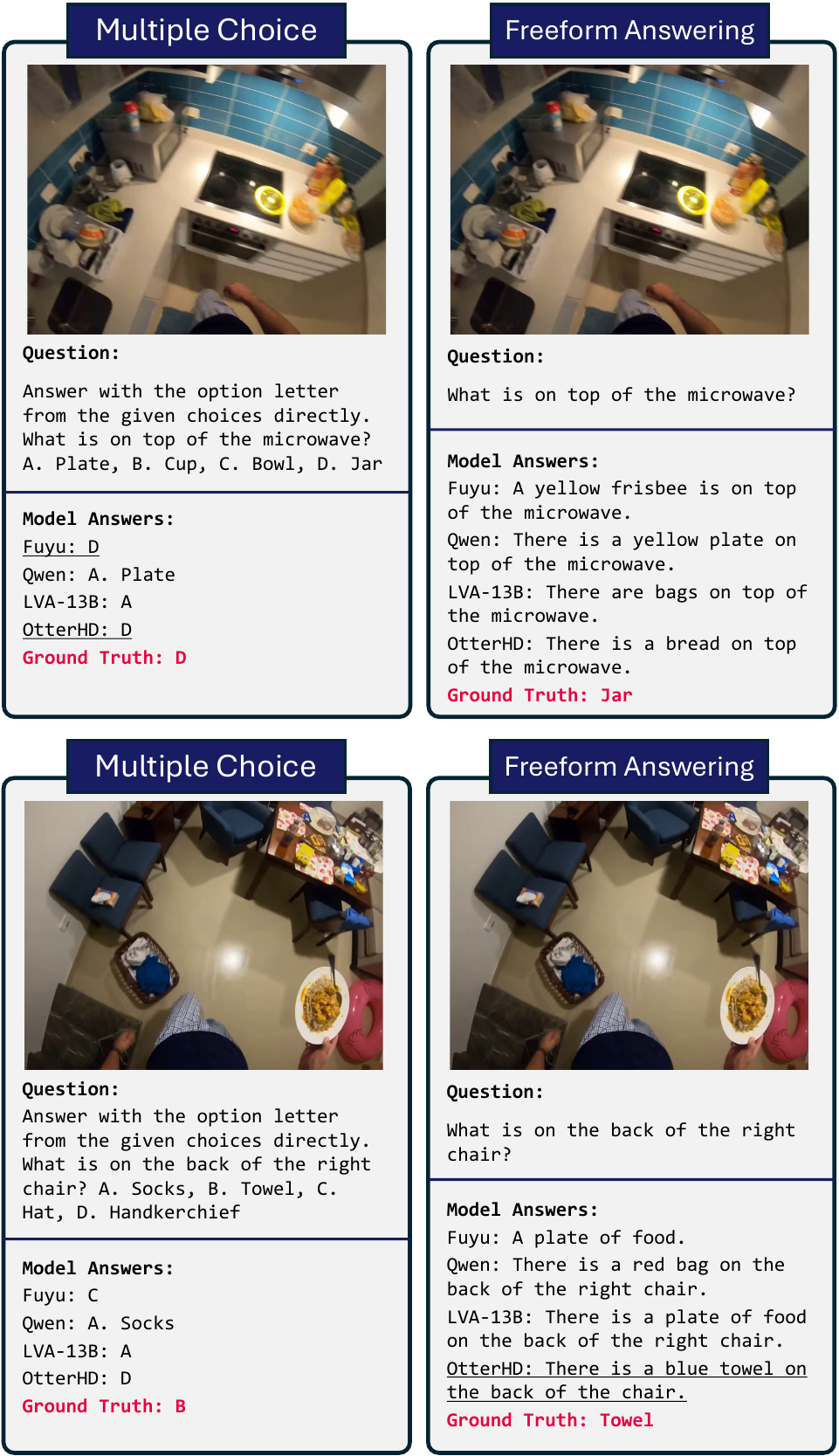}
    \caption{Comparison of different LLMs responses on MagnifierBench on two types of questions. The correct responses are underlined.}
    \label{fig:zib_demo}
\end{figure*}

In Figure \ref{fig:zib_demo}, we present a comparative analysis of various LMMs when evaluated on the Magnifier Benchmark, which encompasses two primary question types: Multiple Choice and Freeform Answering. On the left panel, we observe the multiple-choice questions where models are expected to choose the correct option based on the visual content. On the right panel, freeform answering questions require models to generate a textual description corresponding to the visual query. The ground truth highlights the desired responses. 

A noticeable observation is the variability in the answers provided by different models, emphasizing the complexity and challenges associated with integrating visual and textual information. From the above example provided in Figure \ref{fig:zib_demo}, it is evident that when options are given, i.e., evaluated with Multiple Choice, both Fuyu and OtterHD selected the correct answer D. However, when no options were provided as hints, their answers were incorrect. Consequently, these two evaluation methods can be used to verify the different behaviors of models on the MagnifierBench. Moreover, we have open-sourced all the question and answer logs of the models for further analysis by the community.

\clearpage
\bibliography{ref}

\begin{thebibliography}{10}

\bibitem{flamingo}
Jean-Baptiste Alayrac, Jeff Donahue, Pauline Luc, Antoine Miech, Iain Barr, Yana Hasson, Karel Lenc, Arthur Mensch, Katherine Millican, Malcolm Reynolds, et~al.
\newblock Flamingo: a visual language model for few-shot learning.
\newblock {\em Advances in Neural Information Processing Systems}, 35:23716--23736, 2022.

\bibitem{antol2015vqa}
Stanislaw Antol, Aishwarya Agrawal, Jiasen Lu, Margaret Mitchell, Dhruv Batra, C~Lawrence Zitnick, and Devi Parikh.
\newblock Vqa: Visual question answering.
\newblock In {\em Proceedings of the IEEE international conference on computer vision}, pages 2425--2433, 2015.

\bibitem{open_flamingo}
Anas Awadalla, Irena Gao, Joshua Gardner, Jack Hessel, Yusuf Hanafy, Wanrong Zhu, Kalyani Marathe, Yonatan Bitton, Samir Gadre, Jenia Jitsev, Simon Kornblith, Pang~Wei Koh, Gabriel Ilharco, Mitchell Wortsman, and Ludwig Schmidt.
\newblock Openflamingo, March 2023.

\bibitem{Qwen-VL}
Jinze Bai, Shuai Bai, Shusheng Yang, Shijie Wang, Sinan Tan, Peng Wang, Junyang Lin, Chang Zhou, and Jingren Zhou.
\newblock Qwen-vl: A versatile vision-language model for understanding, localization, text reading, and beyond.
\newblock {\em arXiv preprint arXiv:2308.12966}, 2023.

\bibitem{fuyu-8b}
Rohan Bavishi, Erich Elsen, Curtis Hawthorne, Maxwell Nye, Augustus Odena, Arushi Somani, and Sa\u{g}nak Ta\c{s}\i{}rlar.
\newblock Introducing our multimodal models, 2023.

\bibitem{gpt3}
Tom Brown, Benjamin Mann, Nick Ryder, Melanie Subbiah, Jared~D Kaplan, Prafulla Dhariwal, Arvind Neelakantan, Pranav Shyam, Girish Sastry, Amanda Askell, et~al.
\newblock Language models are few-shot learners.
\newblock {\em Advances in neural information processing systems}, 33:1877--1901, 2020.

\bibitem{chen2023pali}
Xi~Chen, Josip Djolonga, Piotr Padlewski, Basil Mustafa, Soravit Changpinyo, Jialin Wu, Carlos~Riquelme Ruiz, Sebastian Goodman, Xiao Wang, Yi~Tay, et~al.
\newblock Pali-x: On scaling up a multilingual vision and language model.
\newblock {\em arXiv preprint arXiv:2305.18565}, 2023.

\bibitem{chen2023pali3}
Xi~Chen, Xiao Wang, Lucas Beyer, Alexander Kolesnikov, Jialin Wu, Paul Voigtlaender, Basil Mustafa, Sebastian Goodman, Ibrahim Alabdulmohsin, Piotr Padlewski, Daniel Salz, Xi~Xiong, Daniel Vlasic, Filip Pavetic, Keran Rong, Tianli Yu, Daniel Keysers, Xiaohua Zhai, and Radu Soricut.
\newblock Pali-3 vision language models: Smaller, faster, stronger, 2023.

\bibitem{chen2022pali}
Xi~Chen, Xiao Wang, Soravit Changpinyo, AJ~Piergiovanni, Piotr Padlewski, Daniel Salz, Sebastian Goodman, Adam Grycner, Basil Mustafa, Lucas Beyer, et~al.
\newblock Pali: A jointly-scaled multilingual language-image model.
\newblock {\em arXiv preprint arXiv:2209.06794}, 2022.

\bibitem{chen2015microsoft}
Xinlei Chen, Hao Fang, Tsung-Yi Lin, Ramakrishna Vedantam, Saurabh Gupta, Piotr Doll{\'a}r, and C~Lawrence Zitnick.
\newblock Microsoft coco captions: Data collection and evaluation server.
\newblock {\em arXiv preprint arXiv:1504.00325}, 2015.

\bibitem{vicuna2023}
Wei-Lin Chiang, Zhuohan Li, Zi~Lin, Ying Sheng, Zhanghao Wu, Hao Zhang, Lianmin Zheng, Siyuan Zhuang, Yonghao Zhuang, Joseph~E. Gonzalez, Ion Stoica, and Eric~P. Xing.
\newblock Vicuna: An open-source chatbot impressing gpt-4 with 90\%* chatgpt quality, March 2023.

\bibitem{chowdhery2022palm}
Aakanksha Chowdhery, Sharan Narang, Jacob Devlin, Maarten Bosma, Gaurav Mishra, Adam Roberts, Paul Barham, Hyung~Won Chung, Charles Sutton, Sebastian Gehrmann, Parker Schuh, Kensen Shi, Sasha Tsvyashchenko, Joshua Maynez, Abhishek Rao, Parker Barnes, Yi~Tay, Noam Shazeer, Vinodkumar Prabhakaran, Emily Reif, Nan Du, Ben Hutchinson, Reiner Pope, James Bradbury, Jacob Austin, Michael Isard, Guy Gur-Ari, Pengcheng Yin, Toju Duke, Anselm Levskaya, Sanjay Ghemawat, Sunipa Dev, Henryk Michalewski, Xavier Garcia, Vedant Misra, Kevin Robinson, Liam Fedus, Denny Zhou, Daphne Ippolito, David Luan, Hyeontaek Lim, Barret Zoph, Alexander Spiridonov, Ryan Sepassi, David Dohan, Shivani Agrawal, Mark Omernick, Andrew~M. Dai, Thanumalayan~Sankaranarayana Pillai, Marie Pellat, Aitor Lewkowycz, Erica Moreira, Rewon Child, Oleksandr Polozov, Katherine Lee, Zongwei Zhou, Xuezhi Wang, Brennan Saeta, Mark Diaz, Orhan Firat, Michele Catasta, Jason Wei, Kathy Meier-Hellstern, Douglas Eck, Jeff Dean, Slav Petrov, and Noah Fiedel.
\newblock Palm: Scaling language modeling with pathways, 2022.

\bibitem{dai2023instructblip}
Wenliang Dai, Junnan Li, Dongxu Li, Anthony Meng~Huat Tiong, Junqi Zhao, Weisheng Wang, Boyang Li, Pascale Fung, and Steven Hoi.
\newblock Instructblip: Towards general-purpose vision-language models with instruction tuning, 2023.

\bibitem{epic}
Dima Damen, Hazel Doughty, Giovanni~Maria Farinella, Sanja Fidler, Antonino Furnari, Evangelos Kazakos, Davide Moltisanti, Jonathan Munro, Toby Perrett, Will Price, et~al.
\newblock The epic-kitchens dataset: Collection, challenges and baselines.
\newblock {\em IEEE Transactions on Pattern Analysis and Machine Intelligence}, 43(11):4125--4141, 2020.

\bibitem{dao2023flashattention2}
Tri Dao.
\newblock Flash{A}ttention-2: Faster attention with better parallelism and work partitioning.
\newblock 2023.

\bibitem{dehghani2023scaling}
Mostafa Dehghani, Josip Djolonga, Basil Mustafa, Piotr Padlewski, Jonathan Heek, Justin Gilmer, Andreas Steiner, Mathilde Caron, Robert Geirhos, Ibrahim Alabdulmohsin, Rodolphe Jenatton, Lucas Beyer, Michael Tschannen, Anurag Arnab, Xiao Wang, Carlos Riquelme, Matthias Minderer, Joan Puigcerver, Utku Evci, Manoj Kumar, Sjoerd van Steenkiste, Gamaleldin~F. Elsayed, Aravindh Mahendran, Fisher Yu, Avital Oliver, Fantine Huot, Jasmijn Bastings, Mark~Patrick Collier, Alexey Gritsenko, Vighnesh Birodkar, Cristina Vasconcelos, Yi~Tay, Thomas Mensink, Alexander Kolesnikov, Filip Pavetić, Dustin Tran, Thomas Kipf, Mario Lučić, Xiaohua Zhai, Daniel Keysers, Jeremiah Harmsen, and Neil Houlsby.
\newblock Scaling vision transformers to 22 billion parameters, 2023.

\bibitem{deng2009imagenet}
Jia Deng, Wei Dong, Richard Socher, Li-Jia Li, Kai Li, and Li~Fei-Fei.
\newblock Imagenet: A large-scale hierarchical image database.
\newblock In {\em 2009 IEEE conference on computer vision and pattern recognition}, pages 248--255. Ieee, 2009.

\bibitem{dosovitskiy2021image}
Alexey Dosovitskiy, Lucas Beyer, Alexander Kolesnikov, Dirk Weissenborn, Xiaohua Zhai, Thomas Unterthiner, Mostafa Dehghani, Matthias Minderer, Georg Heigold, Sylvain Gelly, Jakob Uszkoreit, and Neil Houlsby.
\newblock An image is worth 16x16 words: Transformers for image recognition at scale, 2021.

\bibitem{persimmon-8b}
Erich Elsen, Augustus Odena, Maxwell Nye, Sa\u{g}nak Ta\c{s}\i{}rlar, Tri Dao, Curtis Hawthorne, Deepak Moparthi, and Arushi Somani.
\newblock Releasing {Persimmon-8B}, 2023.

\bibitem{fang2023eva}
Yuxin Fang, Wen Wang, Binhui Xie, Quan Sun, Ledell Wu, Xinggang Wang, Tiejun Huang, Xinlong Wang, and Yue Cao.
\newblock Eva: Exploring the limits of masked visual representation learning at scale.
\newblock In {\em Proceedings of the IEEE/CVF Conference on Computer Vision and Pattern Recognition}, pages 19358--19369, 2023.

\bibitem{fu2023mme}
Chaoyou Fu, Peixian Chen, Yunhang Shen, Yulei Qin, Mengdan Zhang, Xu~Lin, Zhenyu Qiu, Wei Lin, Jinrui Yang, Xiawu Zheng, et~al.
\newblock Mme: A comprehensive evaluation benchmark for multimodal large language models.
\newblock {\em arXiv preprint arXiv:2306.13394}, 2023.

\bibitem{ego4d}
Kristen Grauman, Andrew Westbury, Eugene Byrne, Zachary Chavis, Antonino Furnari, Rohit Girdhar, Jackson Hamburger, Hao Jiang, Miao Liu, Xingyu Liu, et~al.
\newblock Ego4d: Around the world in 3,000 hours of egocentric video.
\newblock In {\em Proceedings of the IEEE/CVF Conference on Computer Vision and Pattern Recognition}, pages 18995--19012, 2022.

\bibitem{hudson2019gqa}
Drew~A Hudson and Christopher~D Manning.
\newblock Gqa: A new dataset for real-world visual reasoning and compositional question answering.
\newblock In {\em Proceedings of the IEEE/CVF conference on computer vision and pattern recognition}, pages 6700--6709, 2019.

\bibitem{laurenccon2023obelisc}
Hugo Lauren{\c{c}}on, Lucile Saulnier, L{\'e}o Tronchon, Stas Bekman, Amanpreet Singh, Anton Lozhkov, Thomas Wang, Siddharth Karamcheti, Alexander~M Rush, Douwe Kiela, et~al.
\newblock Obelisc: An open web-scale filtered dataset of interleaved image-text documents.
\newblock {\em arXiv preprint arXiv:2306.16527}, 2023.

\bibitem{li2023otter}
Bo~Li, Yuanhan Zhang, Liangyu Chen, Jinghao Wang, Jingkang Yang, and Ziwei Liu.
\newblock Otter: A multi-modal model with in-context instruction tuning, 2023.

\bibitem{li2023seedbench}
Bohao Li, Rui Wang, Guangzhi Wang, Yuying Ge, Yixiao Ge, and Ying Shan.
\newblock Seed-bench: Benchmarking multimodal llms with generative comprehension, 2023.

\bibitem{li2023blip}
Junnan Li, Dongxu Li, Silvio Savarese, and Steven Hoi.
\newblock Blip-2: Bootstrapping language-image pre-training with frozen image encoders and large language models.
\newblock {\em arXiv preprint arXiv:2301.12597}, 2023.

\bibitem{li2022blip}
Junnan Li, Dongxu Li, Caiming Xiong, and Steven Hoi.
\newblock Blip: Bootstrapping language-image pre-training for unified vision-language understanding and generation.
\newblock In {\em International Conference on Machine Learning}, pages 12888--12900. PMLR, 2022.

\bibitem{li2023evaluating}
Yifan Li, Yifan Du, Kun Zhou, Jinpeng Wang, Wayne~Xin Zhao, and Ji-Rong Wen.
\newblock Evaluating object hallucination in large vision-language models.
\newblock {\em arXiv preprint arXiv:2305.10355}, 2023.

\bibitem{liu2023improved}
Haotian Liu, Chunyuan Li, Yuheng Li, and Yong~Jae Lee.
\newblock Improved baselines with visual instruction tuning.
\newblock {\em arXiv preprint arXiv:2310.03744}, 2023.

\bibitem{llava}
Haotian Liu, Chunyuan Li, Qingyang Wu, and Yong~Jae Lee.
\newblock Visual instruction tuning.
\newblock {\em arXiv preprint arXiv:2304.08485}, 2023.

\bibitem{liu2021survey}
Yang Liu, Peng Sun, Nickolas Wergeles, and Yi~Shang.
\newblock A survey and performance evaluation of deep learning methods for small object detection.
\newblock {\em Expert Systems with Applications}, 172:114602, 2021.

\bibitem{liu2022open}
Yen-Cheng Liu, Chih-Yao Ma, Xiaoliang Dai, Junjiao Tian, Peter Vajda, Zijian He, and Zsolt Kira.
\newblock Open-set semi-supervised object detection.
\newblock In {\em European Conference on Computer Vision}, pages 143--159. Springer, 2022.

\bibitem{liu2023mmbench}
Yuan Liu, Haodong Duan, Yuanhan Zhang, Bo~Li, Songyang Zhang, Wangbo Zhao, Yike Yuan, Jiaqi Wang, Conghui He, Ziwei Liu, et~al.
\newblock Mmbench: Is your multi-modal model an all-around player?
\newblock {\em arXiv preprint arXiv:2307.06281}, 2023.

\bibitem{lu2023mathvista}
Pan Lu, Hritik Bansal, Tony Xia, Jiacheng Liu, Chunyuan Li, Hannaneh Hajishirzi, Hao Cheng, Kai-Wei Chang, Michel Galley, and Jianfeng Gao.
\newblock Mathvista: Evaluating mathematical reasoning of foundation models in visual contexts.
\newblock {\em arXiv preprint arXiv:2310.02255}, 2023.

\bibitem{marino2019ok}
Kenneth Marino, Mohammad Rastegari, Ali Farhadi, and Roozbeh Mottaghi.
\newblock Ok-vqa: A visual question answering benchmark requiring external knowledge.
\newblock In {\em Proceedings of the IEEE/cvf conference on computer vision and pattern recognition}, pages 3195--3204, 2019.

\bibitem{mathew2021docvqa}
Minesh Mathew, Dimosthenis Karatzas, and C.~V. Jawahar.
\newblock Docvqa: A dataset for vqa on document images, 2021.

\bibitem{mishraICDAR19}
Anand Mishra, Shashank Shekhar, Ajeet~Kumar Singh, and Anirban Chakraborty.
\newblock Ocr-vqa: Visual question answering by reading text in images.
\newblock In {\em ICDAR}, 2019.

\bibitem{mori1999optical}
Shunji Mori, Hirobumi Nishida, and Hiromitsu Yamada.
\newblock {\em Optical character recognition}.
\newblock John Wiley \& Sons, Inc., 1999.

\bibitem{gpt4}
OpenAI.
\newblock Gpt-4 technical report.
\newblock 2023.

\bibitem{chatgpt}
OpenAI.
\newblock Introducing chatgpt.
\newblock 2023.

\bibitem{clip}
Alec Radford, Jong~Wook Kim, Chris Hallacy, Aditya Ramesh, Gabriel Goh, Sandhini Agarwal, Girish Sastry, Amanda Askell, Pamela Mishkin, Jack Clark, et~al.
\newblock Learning transferable visual models from natural language supervision.
\newblock In {\em International conference on machine learning}, pages 8748--8763. PMLR, 2021.

\bibitem{radford2021learning}
Alec Radford, Jong~Wook Kim, Chris Hallacy, Aditya Ramesh, Gabriel Goh, Sandhini Agarwal, Girish Sastry, Amanda Askell, Pamela Mishkin, Jack Clark, et~al.
\newblock Learning transferable visual models from natural language supervision.
\newblock In {\em International conference on machine learning}, pages 8748--8763. PMLR, 2021.

\bibitem{raffel2023exploring}
Colin Raffel, Noam Shazeer, Adam Roberts, Katherine Lee, Sharan Narang, Michael Matena, Yanqi Zhou, Wei Li, and Peter~J. Liu.
\newblock Exploring the limits of transfer learning with a unified text-to-text transformer, 2023.

\bibitem{schwenk2022okvqa}
Dustin Schwenk, Apoorv Khandelwal, Christopher Clark, Kenneth Marino, and Roozbeh Mottaghi.
\newblock A-okvqa: A benchmark for visual question answering using world knowledge.
\newblock In {\em European Conference on Computer Vision}, pages 146--162. Springer, 2022.

\bibitem{vidor}
Xindi Shang, Donglin Di, Junbin Xiao, Yu~Cao, Xun Yang, and Tat-Seng Chua.
\newblock Annotating objects and relations in user-generated videos.
\newblock In {\em Proceedings of the 2019 on International Conference on Multimedia Retrieval}, pages 279--287, 2019.

\bibitem{sidorov2020textcaps}
Oleksii Sidorov, Ronghang Hu, Marcus Rohrbach, and Amanpreet Singh.
\newblock Textcaps: a dataset for image captioning with reading comprehension, 2020.

\bibitem{singh2019towards}
Amanpreet Singh, Vivek Natarajan, Meet Shah, Yu~Jiang, Xinlei Chen, Dhruv Batra, Devi Parikh, and Marcus Rohrbach.
\newblock Towards vqa models that can read.
\newblock In {\em Proceedings of the IEEE/CVF conference on computer vision and pattern recognition}, pages 8317--8326, 2019.

\bibitem{so2022primer}
David~R. So, Wojciech Mańke, Hanxiao Liu, Zihang Dai, Noam Shazeer, and Quoc~V. Le.
\newblock Primer: Searching for efficient transformers for language modeling, 2022.

\bibitem{su2022roformer}
Jianlin Su, Yu~Lu, Shengfeng Pan, Ahmed Murtadha, Bo~Wen, and Yunfeng Liu.
\newblock Roformer: Enhanced transformer with rotary position embedding, 2022.

\bibitem{sun2023aligning}
Zhiqing Sun, Sheng Shen, Shengcao Cao, Haotian Liu, Chunyuan Li, Yikang Shen, Chuang Gan, Liang-Yan Gui, Yu-Xiong Wang, Yiming Yang, et~al.
\newblock Aligning large multimodal models with factually augmented rlhf.
\newblock {\em arXiv preprint arXiv:2309.14525}, 2023.

\bibitem{alpaca}
Rohan Taori, Ishaan Gulrajani, Tianyi Zhang, Yann Dubois, Xuechen Li, Carlos Guestrin, Percy Liang, and Tatsunori~B. Hashimoto.
\newblock Stanford alpaca: An instruction-following llama model.
\newblock \url{https://github.com/tatsu-lab/stanford_alpaca}, 2023.

\bibitem{tong2020recent}
Kang Tong, Yiquan Wu, and Fei Zhou.
\newblock Recent advances in small object detection based on deep learning: A review.
\newblock {\em Image and Vision Computing}, 97:103910, 2020.

\bibitem{llama}
Hugo Touvron, Thibaut Lavril, Gautier Izacard, Xavier Martinet, Marie-Anne Lachaux, Timoth{\'e}e Lacroix, Baptiste Rozi{\`e}re, Naman Goyal, Eric Hambro, Faisal Azhar, Aurelien Rodriguez, Armand Joulin, Edouard Grave, and Guillaume Lample.
\newblock Llama: Open and efficient foundation language models.
\newblock {\em arXiv preprint arXiv:2302.13971}, 2023.

\bibitem{touvron2023llama}
Hugo Touvron, Thibaut Lavril, Gautier Izacard, Xavier Martinet, Marie-Anne Lachaux, Timothée Lacroix, Baptiste Rozière, Naman Goyal, Eric Hambro, Faisal Azhar, Aurelien Rodriguez, Armand Joulin, Edouard Grave, and Guillaume Lample.
\newblock Llama: Open and efficient foundation language models, 2023.

\bibitem{wolf-etal-2020-transformers}
Thomas Wolf, Lysandre Debut, Victor Sanh, Julien Chaumond, Clement Delangue, Anthony Moi, Pierric Cistac, Tim Rault, Rémi Louf, Morgan Funtowicz, Joe Davison, Sam Shleifer, Patrick von Platen, Clara Ma, Yacine Jernite, Julien Plu, Canwen Xu, Teven~Le Scao, Sylvain Gugger, Mariama Drame, Quentin Lhoest, and Alexander~M. Rush.
\newblock Transformers: State-of-the-art natural language processing.
\newblock In {\em Proceedings of the 2020 Conference on Empirical Methods in Natural Language Processing: System Demonstrations}, pages 38--45, Online, October 2020. Association for Computational Linguistics.

\bibitem{yang2023panoptic}
Jingkang Yang, Wenxuan Peng, Xiangtai Li, Zujin Guo, Liangyu Chen, Bo~Li, Zheng Ma, Kaiyang Zhou, Wayne Zhang, Chen~Change Loy, et~al.
\newblock Panoptic video scene graph generation.
\newblock In {\em Proceedings of the IEEE/CVF Conference on Computer Vision and Pattern Recognition}, pages 18675--18685, 2023.

\bibitem{yu2016modeling}
Licheng Yu, Patrick Poirson, Shan Yang, Alexander~C Berg, and Tamara~L Berg.
\newblock Modeling context in referring expressions.
\newblock In {\em Computer Vision--ECCV 2016: 14th European Conference, Amsterdam, The Netherlands, October 11-14, 2016, Proceedings, Part II 14}, pages 69--85. Springer, 2016.

\bibitem{yu2023mm}
Weihao Yu, Zhengyuan Yang, Linjie Li, Jianfeng Wang, Kevin Lin, Zicheng Liu, Xinchao Wang, and Lijuan Wang.
\newblock Mm-vet: Evaluating large multimodal models for integrated capabilities.
\newblock {\em arXiv preprint arXiv:2308.02490}, 2023.

\bibitem{zhang2023transfer}
Ao~Zhang, Hao Fei, Yuan Yao, Wei Ji, Li~Li, Zhiyuan Liu, and Tat-Seng Chua.
\newblock Transfer visual prompt generator across llms.
\newblock {\em arXiv preprint arXiv:2305.01278}, 2023.

\end{thebibliography}
\bibliographystyle{plain}
\end{document}